\documentclass[11pt]{article}

\usepackage[final]{acl}

\usepackage{times}
\usepackage{latexsym}
\usepackage{enumitem}
\usepackage[T1]{fontenc}

\usepackage[utf8]{inputenc}
\usepackage{textcomp}
\usepackage{microtype}

\usepackage{inconsolata}

\usepackage{graphicx}

\usepackage[nolist]{acronym}

\usepackage{booktabs}
\usepackage{multicol}
\usepackage{multirow}
\usepackage{url}
\usepackage[table]{xcolor}
\definecolor{sectiongray}{HTML}{EFEFEF}
\definecolor{blockgray}{HTML}{F7F7F7}
\usepackage{amsmath}
\usepackage{seqsplit}

\usepackage{tcolorbox}
\tcbuselibrary{breakable, skins}

\newtcolorbox{promptbox}[1][]{%
    breakable,
    enhanced,
    colback=gray!5,
    colframe=gray!50,
    coltitle=black,
    fonttitle=\bfseries\small,
    title=#1,
    boxrule=0.5pt,
    arc=2pt,
    left=6pt, right=6pt, top=4pt, bottom=4pt,
    fontupper=\small\ttfamily,
}

\title{\textsc{ComplexityMT}: Benchmarking the Interaction Between Text Complexity and Machine Translation}

\author{
  \textbf{Joseph Marvin Imperial}\textnormal{\textsuperscript{1,3}},
  \textbf{Junhong Liang}\textnormal{\textsuperscript{4}},
  \textbf{Belal Shoer}\textnormal{\textsuperscript{4}},
  \textbf{Abdullah Barayan}\textnormal{\textsuperscript{2,10}},\\
  \textbf{Rodrigo Wilkens}\textsuperscript{5},
  \textbf{Omar Mussa}\textsuperscript{11},
  \textbf{Dawn Knight}\textsuperscript{2},
  \textbf{Eugénio Ribeiro}\textsuperscript{8,9},\\
  \textbf{Ekaterina Kochmar}\textsuperscript{4,6},
  \textbf{Sowmya Vajjala}\textsuperscript{7},
  \textbf{Fernando Alva-Manchego}\textsuperscript{2},\\
  \textbf{Harish Tayyar Madabushi}\textsuperscript{1}
\\
  \textsuperscript{1}University of Bath,
  \textsuperscript{2}Cardiff University,
  \textsuperscript{3}National University Philippines,
  \textsuperscript{4}MBZUAI,\\
  \textsuperscript{5}University of Exeter,
  \textsuperscript{6}University of Victoria,
  \textsuperscript{7}National Research Council, Canada,\\
  \textsuperscript{8}INESC-ID Lisboa
  \textsuperscript{9}Instituto Universitário de Lisboa (ISCTE-IUL), ISTAR\\
  \textsuperscript{10}King Abdulaziz University,
  \textsuperscript{11}Saudi Electronic University\\
  \small{
    \texttt{\href{mailto:jmri20@bath.ac.uk}{jmri20@bath.ac.uk}},
    \texttt{\href{mailto:htm43@bath.ac.uk}{htm43@bath.ac.uk}},
    \texttt{\href{mailto:alvamanchegof@cardiff.ac.uk}{alvamanchegof@cardiff.ac.uk}}
  }
}

\begin{document}

\begin{acronym}
    \acro{CEFR}{Common European Framework of Reference for Languages}
    \acro{LLM}{Large Language Model}
    \acro{MT}{Machine Translation}
    \acro{NLP}{Natural Language Processing}
\end{acronym}

\maketitle
\begin{abstract}

\textit{When a text is translated, does the translation retain the complexity of the original?} 
We introduce \textbf{\textsc{ComplexityMT}}, a new challenge for assessing how text complexity and machine translation interact with and influence each other, using the 
\ac{CEFR} levels as the measure of text complexity. Across six languages, including Arabic, Dutch, English, French, Hindi, and Russian, we evaluate three open-weight models, one closed model, and a commercial machine translation system on two tasks: i) \textit{correlation} of \ac{CEFR} with translation difficulty, and ii) \textit{shifts} in \ac{CEFR} levels of the source texts. Our experiments show that higher \ac{CEFR} levels make texts more difficult to translate and that machine translation lowers the target text's CEFR level for most languages at the document level. These findings provide new insights for researchers and practitioners working on multilingual pedagogical content generation and MT difficulty estimation.
\end{abstract}

\begin{tabular}{ll}
\raisebox{-0.3em}{\includegraphics[height=1em]{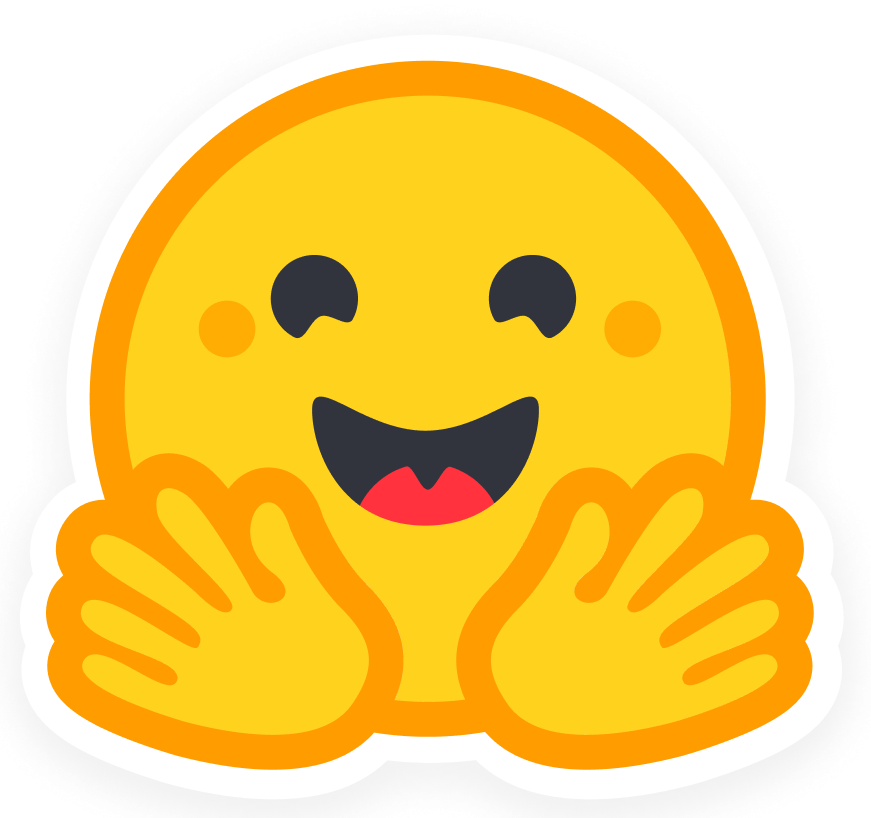}}  & \href{https://huggingface.co/datasets/UniversalCEFR/ComplexityMT}{\texttt{UniversalCEFR/ComplexityMT}} \\
\raisebox{-0.3em}{\includegraphics[height=1em]{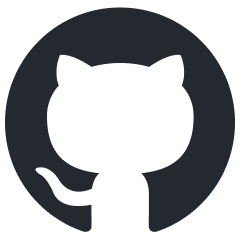}} & \href{https://github.com/UniversalCEFR/ComplexityMT}{\texttt{UniversalCEFR/ComplexityMT}} \\
\end{tabular}

\begin{figure}[!t]
    \centering
    \includegraphics[width=1\linewidth]{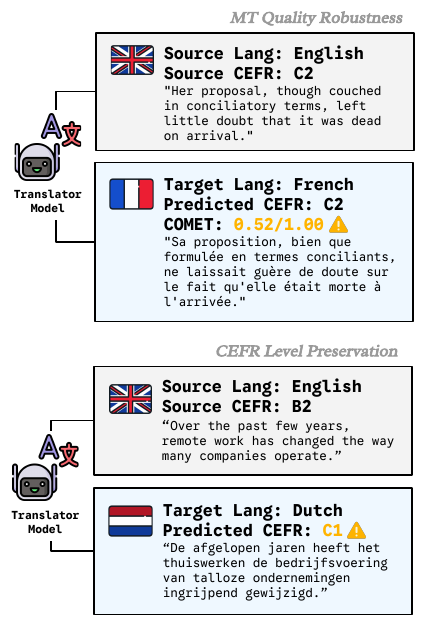}
    \caption{We observe two limitations in existing machine translation models: i) a \textit{robustness} problem where they tend to produce lower quality translations (via COMET or GEMBA) for higher-complexity source texts, and ii) a \textit{preservation} problem where they tend to change the pedagogical complexity levels (e.g., CEFR) of the source text.}
    \label{fig:pedagogical_preservation_task}
\end{figure}

\section{Introduction}
\label{sec:intro}

Creating reading materials tailored to a learner's proficiency level has been a topic of interest in educational research for a long time. Approaches such as traditional readability formulas \cite{Kincaid1975,DuBay2004,Crossley2017} and proprietary measures such as Lexile scores \cite{lennon2004lexile} were used in the past to evaluate existing materials and adapt them to specific reading levels. More recently, \ac{NLP} research has enabled automation of two foundational tasks in this direction, namely automatic readability assessment \cite{aluisio-etal-2010-readability,ciobanu-etal-2015-readability,xia-etal-2016-text,deutsch-etal-2020-linguistic,vajjala-2022-trends} and text simplification \cite{maddela-xu-2018-word,scarton-specia-2018-learning,nishihara-etal-2019-controllable,alva-manchego-etal-2020-data,maddela-etal-2021-controllable,sheang-saggion-2021-controllable,alva-manchego-etal-2025-findings,barayan-etal-2025-analysing}, which could together support the creation of targeted reading level-specific texts. Research in this direction has understandably focused on high-resource languages like English due to the abundance of data and models. \ac{MT} 
offers a natural method to extend reading level-appropriate texts into more languages \cite{xu-etal-2016-optimizing,marchisio-etal-2019-controlling,alva-manchego-shardlow-2022-towards,zouhar-etal-2026-generating}. But understanding whether existing MT models are actually fit for this purpose requires empirical evidence of how text complexity affects translation quality, and if it preserves the complexity of the source text being translated. Understanding these interactions will guide how to leverage MT and text simplification to scale pedagogical content generation across languages. 


The assumption that complex text could be difficult to translate is not new. Text simplification was considered a pre-processing step to reduce translation difficulty and improve translation quality, both in earlier MT models as well as more recently \cite[e.g.,][]{chandrasekar-etal-1996-motivations,mehta2020simplify}, and recent work has shown that \textit{complex texts are difficult to translate} \cite{shardlow-alva-manchego-2022-simple}. At the same time, a separate line of work describes \textit{simplification as a translation universal}, i.e., translated text is shown to be more readable and less complex, as it relies on high-frequency tokens in the target language \cite{corpas-pastor-etal-2008-translation,lu-etal-2021-unsupervised-method,wastl-etal-2025-machine}. In all these cases, the notion of complexity has predominantly been binary. Drawing on what appears to be contrasting viewpoints on the relationship between text complexity and machine translation, we revisit these questions in our paper, specifically in the context of pedagogical difficulty, which we operationalize using the \ac{CEFR} scale \cite{CouncilEurope2001}. Thus, we pose the following research questions:

\begin{itemize}[leftmargin=1.5em]
    \item RQ1: How does the \ac{CEFR} level of a text \textit{correlate} with translation difficulty?
    \item RQ2: How does the \ac{CEFR} level of a text \textit{change} under translation? 
\end{itemize}

These questions, motivated by problems in multilingual, pedagogically grounded content generation, also contribute to \ac{MT} difficulty estimation research. To our knowledge, the interaction between MT and pedagogical difficulty-based content generation has not been studied in past \ac{NLP} research, and this paper introduces a framework to study this through two tasks (see Section~\ref{sec:core_task}). 

\section{Related Work}
\label{sec:relw}
We consider two strands of existing \ac{NLP} research as directly related to our current work: \ac{MT} difficulty estimation, and complexity-controlled \ac{MT}. 

\paragraph{MT Difficulty Estimation } \ac{MT} is one of the core research problems in \ac{NLP}, and the topics of \ac{MT} difficulty estimation, i.e., how difficult a text is for an \ac{MT} system to translate and how to mitigate this effect, have all received some attention in past research. Various text-level features (e.g., text length and lexical diversity) have been explored to model source text difficulty in machine translation \cite{hale2002interaction,mishra-etal-2013-automatically,li-etal-2014-assessing,bugliarello-etal-2020-easier,araghi2024link,proietti-etal-2025-estimating,zouhar-etal-2026-generating}. \citet{proietti-etal-2025-estimating} have recently considered the \ac{CEFR} scale \cite{CouncilEurope2001} as a way to characterize translation difficulty. Text simplification has been explored as a preprocessing step before \ac{MT}, to make the text easier to (machine) translate and improve translation quality \cite{chandrasekar-etal-1996-motivations,mehta2020simplify}. However, to our knowledge, all prior research on \ac{MT} difficulty estimation has focused primarily on sentence-level text and English source data and has been motivated primarily by improving \ac{MT} translation quality, rather than by a pedagogical justification like ours. 

In this paper, we look at \ac{MT} difficulty estimation by considering the \ac{CEFR} scale as the text difficulty measure, and \ac{MT} quality as the difficulty measure for machine translation.  Unlike other works on this topic, we go beyond sentence-level analyses and English and also consider the document-level across multiple languages to explore this question. 

\paragraph{Complexity Controlled MT } Generating translation variants by controlling for aspects such as  politeness \cite{sennrich-etal-2016-controlling}, formality \cite{nadejde-etal-2022-cocoa}, and personalization \cite{mirkin-meunier-2015-personalized} is well-studied in \ac{NLP} research. In this line of research, controlling for text complexity in \ac{MT} \cite{agrawal-carpuat-2019-controlling,marchisio-etal-2019-controlling,tani-etal-2022-benchmark,zouhar-etal-2026-generating} is somewhat related to our second research question, although the specific question of whether translation can preserve the original text complexity is not explored in previous work. Further, this strand of research assumes that the target translation's text complexity level is prespecified. 

In this paper, we explore whether translation preserves the source language's text complexity in the target language. Compared to previous works, we use a pedagogical complexity construct, CEFR, as our main reference for complexity for measuring shifts from MT models. 

\section{\textsc{ComplexityMT}: Benchmarking the Interaction between Text Complexity and Machine Translation} 
\label{sec:core_task}

As discussed in Section~\ref{sec:relw}, previous work has expressed contrasting views on the relationship between text complexity and \ac{MT} difficulty (as measured by MT quality). In that context, we introduce \textsc{ComplexityMT}, a framework for assessing the impact of text complexity on \ac{MT} across two core aspects: robustness and preservation. \textbf{Robustness} captures the expectation that good \ac{MT} models should maintain translation quality across the text-complexity spectrum, thereby addressing RQ1. \textbf{Preservation} builds on the expectation that text complexity is maintained across translations, addressing RQ2. 
The following sections describe  our experimental pipelines to assess both aspects.

\subsection{\textsc{ComplexityMT-Robustness}}
\label{subsec:robustness}
This task assesses whether \ac{MT} quality correlates with the complexity of the source text. Given a set of \ac{CEFR}-labeled source texts and a set of target languages, the robustness of an \ac{MT} system is assessed as follows:

\begin{enumerate}[leftmargin=1.5em]
    \item We translate each source text $x$ with an assigned gold-standard \ac{CEFR} level $\ell_x \in \{A1, A2, B1, B2, C1, C2\}$ into each target language $L_{\text{tgt}}$ using the 
    MT model under evaluation to produce translation $y$;
    \item We then estimate a reference-free \ac{MT} quality score $q(y) \in [0,1]$;
    \item Finally, we compute the Spearman correlation 
    $\rho = \mathrm{corr}(\ell_x, q(y))$ between the source \ac{CEFR} 
    level and the \ac{MT} quality score.
\end{enumerate}

We use the computed correlation $\rho$ between textual complexity and \ac{MT} quality as the main robustness metric, with values closer to zero indicating higher robustness. A significant negative correlation ($\rho < 0$) indicates that the translation quality decreases as the \ac{CEFR} level increases, implying that higher-level texts pose greater challenges for the \ac{MT} system. Conversely, a positive correlation ($\rho > 0$) suggests that higher-level texts receive better quality scores, a less intuitive but theoretically possible outcome.


\definecolor{sectiongray}{gray}{0.92}
\begin{table*}[!t]
\centering
\small
\begin{tabular}{llrrrrrrr}
\toprule
\multirow{2}{*}{\textbf{Source}} &
\multirow{2}{*}{\textbf{Lang}} &
\multirow{2}{*}{\textbf{Size}} &
\multicolumn{6}{c}{\textbf{CEFR Level Breakdown}} \\
\cmidrule(lr){4-9}
& & & \textbf{A1} & \textbf{A2} & \textbf{B1} & \textbf{B2} & \textbf{C1} & \textbf{C2} \\
\midrule

\rowcolor{sectiongray}
\multicolumn{9}{l}{\textit{Document-Level}} \\
\addlinespace[2pt]
\textsc{ELG-CEFR-NL} \cite{breuker2022cefr}
  & NL & 200 & 4 & 21 & 69 & 77 & 22 & 7 \\
\textsc{ELG-CEFR-EN} \cite{breuker2022cefr}
  & EN & 72 & 3 & 13 & 13 & 17 & 17 & 9 \\
\textsc{CambridgeExams} \cite{xia-etal-2016-text}
  & EN & 43 & 0 & 6 & 9 & 7 & 10 & 11 \\
\textsc{Kwiziq} \cite{vasquez-rodriguez-etal-2022-benchmark}
  & FR & 200 & 9 & 15 & 80 & 65 & 31 & 0 \\
\addlinespace[2pt]
\textbf{Total} & -- & \textbf{515}
  & \textbf{16} & \textbf{55} & \textbf{171}
  & \textbf{166} & \textbf{80} & \textbf{27} \\

\midrule
\rowcolor{sectiongray}
\multicolumn{9}{l}{\textit{Sentence-Level}} \\
\addlinespace[2pt]
\multirow[t]{5}{*}{\textsc{ReadMe++} \cite{naous-etal-2024-readme}}
  & FR & 200 & 23 & 42 & 52 & 35 & 32 & 16 \\
  & RU & 200 & 41 & 36 & 52 & 35 & 24 & 12 \\
  & AR & 200 & 12 & 20 & 57 & 66 & 30 & 15 \\
  & HI & 200 & 38 & 34 & 42 & 42 & 28 & 16 \\
  & EN & 45  & 3  & 6  & 10 & 16 & 9  & 1  \\
\textsc{CEFR-SP} \cite{arase-etal-2022-cefr}
  & EN & 155 & 2 & 17 & 51 & 55 & 28 & 2 \\
\addlinespace[2pt]
\textbf{Total} & -- & \textbf{1,000}
  & \textbf{119} & \textbf{155} & \textbf{264}
  & \textbf{249} & \textbf{151} & \textbf{62} \\
\bottomrule
\end{tabular}
\caption{The distribution of multilingual reference texts extracted from the \textsc{UniversalCEFR} test split ($n = 1,515$) we used for this study across the CEFR levels and formats (document- and sentence-level).}
\label{tab:data_distribution}
\end{table*}

\subsection{\textsc{ComplexityMT-Preservation}}
\label{subsec:preservation}
This task assesses whether \ac{MT} systems preserve the \ac{CEFR} levels of the source texts upon translation. Since obtaining manual text-complexity annotations for the translations generated by each evaluated \ac{MT} system is not feasible, we rely on a pretrained multilingual \ac{CEFR} level classifier (see \S\ref{sec:classifiers} for details). To separate translation effects from the CEFR classifier's calibration errors, we evaluate \ac{CEFR} level preservation via a backtranslation procedure\footnote{A more direct test one can visualize is to compare the gold CEFR level of the source text assessed by human experts with a classifier-provided CEFR level of its forward translation. However, we avoid this setup since we cannot efficiently obtain human CEFR assessments for our translations. Comparing shifts between human expert and classifier-provided labels introduces a \textit{label discrepancy} that would need to be mitigated.} with a model-anchored shift, rather than a direct gold-vs-classifier comparison, as described below:

\begin{enumerate}[leftmargin=1.5em]
    \item Given a source text $x$ in language $L_{\text{src}}$ with 
    the gold-standard \ac{CEFR} level $\ell_{\text{gold}}$, we translate $x$ into a pivot language 
    $L_{\text{piv}}$ with an \ac{MT} model, producing 
    the forward translation $y_{\text{fwd}}$;
    \item We then translate $y_{\text{fwd}}$ back into $L_{\text{src}}$ with 
    the same \ac{MT} model, producing the back-translation 
    $y_{\text{back}}$;
    \item Next, we apply the \ac{CEFR} level classifier $f$ independently to each translation output to obtain the forward and backtranslation \ac{CEFR} level predictions:
    $\hat{\ell}_{\text{fwd}} = f(y_{\text{fwd}})$ and 
    $\hat{\ell}_{\text{back}} = f(y_{\text{back}})$;
    \item Finally, we compute the model-anchored \ac{CEFR} level shift:
    \begin{equation}
    \Delta\ell_{\text{model}} = \hat{\ell}_{\text{back}} 
    - \hat{\ell}_{\text{fwd}}.
\end{equation}
\end{enumerate}

We use the computed \ac{CEFR} level shift $\Delta\ell_{\text{model}}$ as the main preservation metric, where a value of $\Delta\ell_{\text{model}} = 0$ signals that the backtranslation process preserved the original \ac{CEFR} level, while a value of $\Delta\ell_{\text{model}} \neq 0$ indicates a net \ac{CEFR} level shift induced by the \ac{MT} model during translation. 
The purpose of anchoring on the classifier's forward prediction is two-fold. First, it minimizes the classifier's disagreement with the gold label of the source text, thereby leaving the effect exclusively to the classifier's predictions. Second, as mentioned earlier, we lack human expert CEFR labels for the translations. We note that the classifier is applied to texts of two different languages on the two legs of backtranslation ($L_{\text{piv}}$ for $y_{\text{fwd}}$ and $L_{\text{src}}$ for $y_{\text{back}}$), thus per-language prediction differences in the classifier are not canceled in $\Delta\ell_{\text{model}}$. We address this by investigating robustness across three structurally different classifiers (Section~\ref{sec:results}). Given this setup, $\Delta\ell_{\text{model}}$ represents the \textit{net effect} of forward and backward translations combined together, which we treat as the core measure of CEFR preservation.




\section{Experimental Setup}
\label{sec:setup}

In this section, we detail an implementation and application of \textsc{ComplexityMT}. The framework can easily be extended to additional languages, used to evaluate other \ac{MT} systems, and improved through advances in automatic \ac{CEFR} level prediction.

\subsection{Data}
\label{sec:data}
We use a subset of \ac{CEFR}-labeled texts from the test split of \textbf{\textsc{UniversalCEFR}} \citep{imperial-etal-2025-universalcefr} at both the sentence and document levels.\footnote{\url{https://huggingface.co/UniversalCEFR}} 
We filtered \textsc{UniversalCEFR} to extract the reference-level texts that are associated with gold-standard CEFR levels as reported in Table~\ref{tab:data_distribution}. The sentence-level data cover multiple source languages, including English, French, Arabic, Hindi, and Russian, while the document-level data include English, French, and Dutch. Each text is associated with an original \ac{CEFR} label from A1 to C2. For analysis, \ac{CEFR} levels are assigned to an ordinal scale, where A1~=~1, A2~=~2, B1~=~3, B2~=~4, C1~=~5 and C2~=~6. Table~\ref{tab:data_distribution} shows the distribution of \ac{CEFR} levels in the document-level and sentence-level subsets. The document-level subset contains 515 instances across three languages, while the sentence-level subset contains 1,000 instances across five languages, with 200 instances per language.


\begin{figure*}[!t]
    \centering
    \includegraphics[width=1\linewidth]{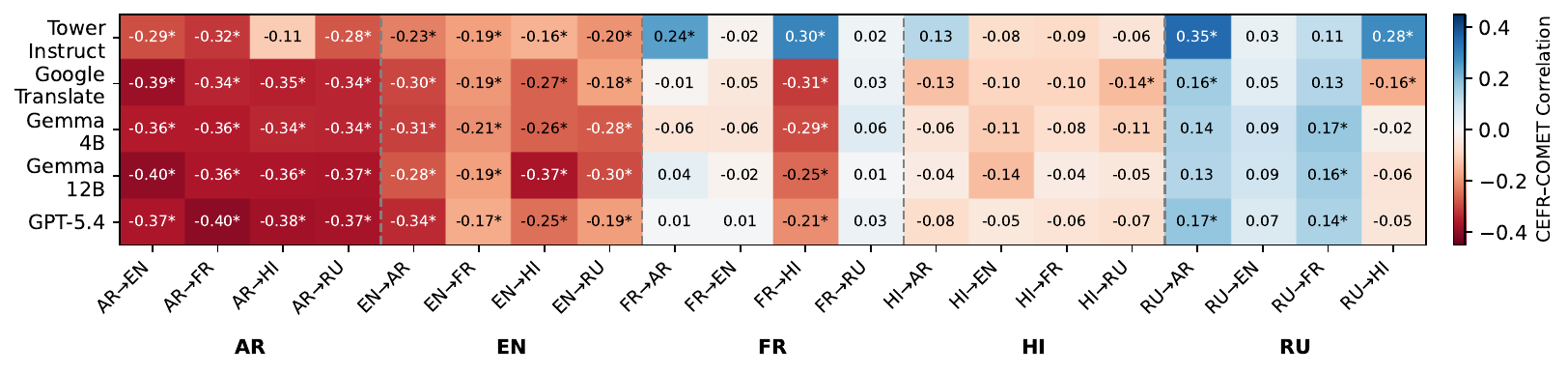}
    \caption{Spearman correlation between source CEFR levels and COMET scores across MT models for sentence-level texts. * indicates statistical significance ($p < 0.05$). Negative values indicate that translation quality, as measured by COMET, decreases as the CEFR level of the source text increases. }
    \label{fig:task1_comet_sentence}
\end{figure*}

\begin{figure*}[!t]
    \centering
    \includegraphics[width=1\linewidth]{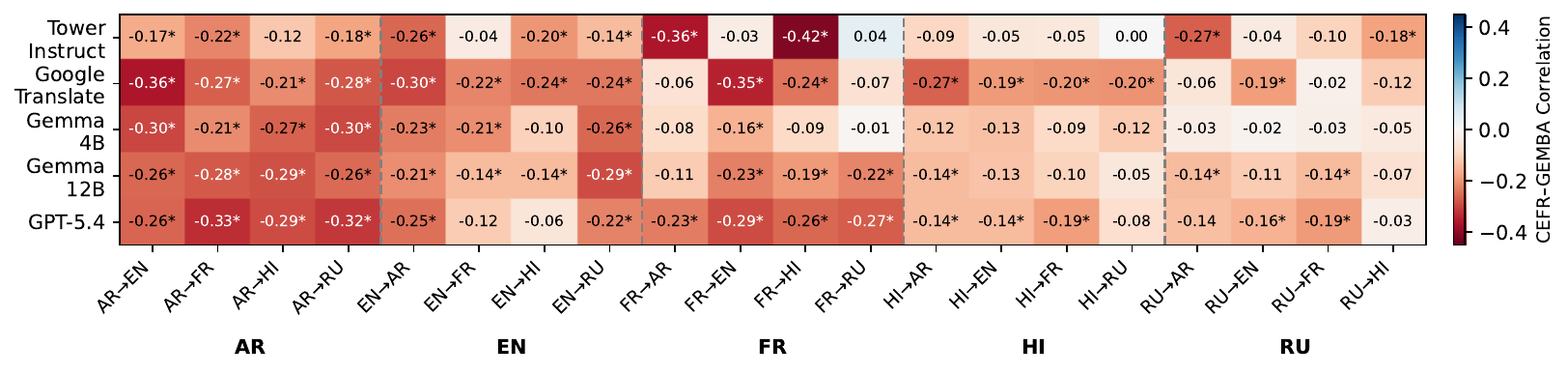}
    \caption{Spearman correlation between source CEFR levels and GEMBA-DA scores across MT models for sentence-level texts. * indicates statistical significance ($p < 0.05$). The significance pattern matches the results of COMET except for Russian where it diverges for GEMBA, although the absolute correlations are smaller.}
    \label{fig:task1_gemba_sentence}
\end{figure*}

\subsection{Translation Models}
\label{sec:models}
We evaluate five diverse \ac{MT} systems spanning different training objectives and scales: a general-purpose LLM {GPT-5.4} \cite{OpenAI2025}; three translation-specialized LLMs, {TowerInstruct-7B} \cite{alves2024tower} and {TranslateGemma} \cite{gemmatranslate2026} in its 4B and 12B variants; and a commercial translation system, {Google Cloud Translation API}.\footnote{\url{https://cloud.google.com/translate}} All models were accessed between February and May 2026. 

\subsection{Translation Quality Metrics}
\label{sec:translation_metrics}
We select two reference-free MT quality metrics for \textsc{ComplexityMT-Robustness}, specifically {COMET}\footnote{\url{https://github.com/Unbabel/COMET}} \cite{rei-etal-2020-comet} which measures quality via an encoder-based multilingual BERT model, and {GEMBA-DA}\footnote{\url{https://github.com/MicrosoftTranslator/GEMBA}} \cite{kocmi-federmann-2023-large}, which measures quality via prompting GPT-5.4 for a direct assessment. Both metrics have been shown to correlate strongly with human evaluation. We use a uniform score scale of $[0,1]$ where higher values indicate better translations.

\begin{figure*}[!t]
    \centering
    \begin{minipage}{0.49\linewidth}
        \centering
        \includegraphics[width=\linewidth]{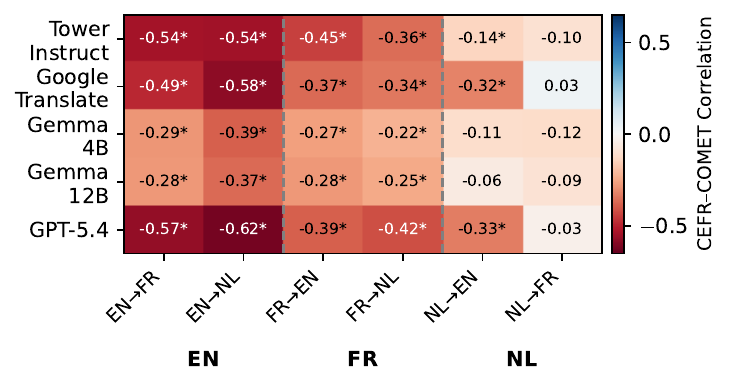}
    \end{minipage}
    \hfill
    \begin{minipage}{0.49\linewidth}
        \centering
        \includegraphics[width=\linewidth]{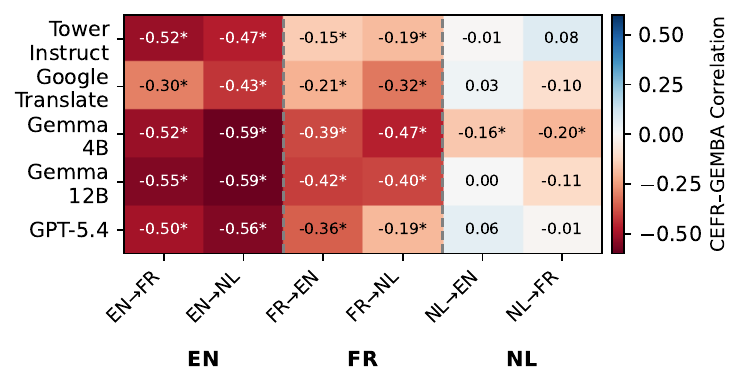}
    \end{minipage}
    \caption{Spearman correlation between source CEFR levels and COMET scores (left) and GEMBA scores (right) across MT models for document-level texts. * indicates statistical significance ($p < 0.05$).}
    \label{fig:task1_document}
\end{figure*}

\begin{figure*}[!t]
    \centering
    \includegraphics[width=1\linewidth]{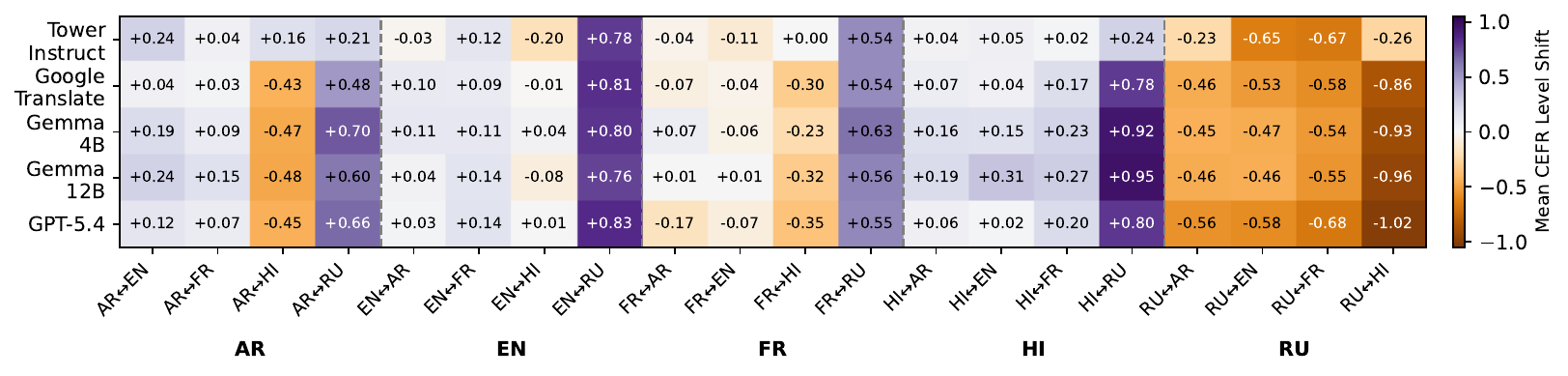}
    \caption{Mean model-anchored CEFR level shifts from back-translations across MT models for sentence-level texts. We anchor the \textsc{UniversalCEFR} classifier's forward prediction to isolate the back-translation's effects from the classifier bias.}
    \label{fig:task2_levelshift_sentence}
\end{figure*}

\subsection{CEFR Level Classifiers}
\label{sec:classifiers}
To assess whether the \ac{CEFR} level classifier impacts the results of the benchmark, we use a recent state-of-the-art {XLM-R} model finetuned with the massively multilingual train split of the \textsc{UniversalCEFR} data. We verified that there were no overlaps between the train split used for training the XLM-R CEFR classifier and the curated test split used for the evaluation for \textsc{ComplexityMT-Robustness} and \textsc{ComplexityMT-Preservation}. Thus, we guarantee that there is no data leakage within the experiments. For comparison, we also perform cross-analysis with two additional CEFR classifiers that were trained with the same data but that differ architecturally -- one is based on the ModernBERT architecture \cite{warner-etal-2025-smarter} and another is trained with 146 hand-engineered linguistic features using Random Forest (see Appendix~\ref{app:classifier_reliability}).

\section{Results}
\label{sec:results}
In this section, we discuss the results of our experiments exploring how text complexity and MT affect each other across languages.

\subsection{Translation Quality Declines with CEFR Level}
We address RQ1 through the \textsc{ComplexityMT-Robustness} task described in Section~\ref{subsec:robustness}. Figures~\ref{fig:task1_comet_sentence} and \ref{fig:task1_gemba_sentence} present Spearman correlation heatmaps capturing the relationship between text complexity and MT quality across the five multilingual MT models
at the sentence level with \textsc{COMET} and \textsc{GEMBA-DA} scores respectively, while Figure~\ref{fig:task1_document} does the same for the document level.

Across most source languages at the sentence level, we observe a distinct pattern where \textit{translation quality correlates negatively with the source CEFR levels} for Arabic and English, meaning that texts of higher CEFR levels are associated with lower translation quality according to COMET scores. These significant correlations are observed within the range of $-0.16$ to $-0.40$, and this pattern is consistent with the general expectation that higher-proficiency texts, which tend to contain more complex syntactic structures, richer vocabulary, and denser information, pose greater challenges for MT models. For GEMBA, we observe the same predominantly negative correlation pattern at smaller magnitudes. Interestingly, Russian stands out as an outlier with positive COMET correlations with CEFR level not seen under GEMBA. We posit that most MT models have been trained on higher-level Russian texts, which are more likely to conform to conventional written registers, thereby yielding higher translation scores.

At the document level, we observe that \textit{negative correlation between translation quality and CEFR levels} is stronger and more uniform. For English and French, all MT models show significant negative correlations with both translation quality metrics where $\rho \approx -0.47$ and $-0.34$ for COMET, and $\rho \approx -0.50$ and $-0.31$ for GEMBA. For NL, this correlation is less pronounced with $\rho \approx -0.13$ and $-0.04$. We posit that the effect is more pronounced for documents than for sentences because longer inputs may carry more linguistic complexity, allowing translation difficulty to converge and yield less noise than for a single sentence.


\begin{table}[t]
\centering
\scriptsize
\definecolor{modelrow}{gray}{0.92}
\caption{Mean model-anchored CEFR level shift under three independently designed CEFR classifiers across MT models and granularities (25\% sample), based on $n=1500$ sentence-level and $n=270$ document-level instances.}
\label{tab:threeway-delta}
\begin{tabular}{lrrr}
\toprule
\bf MT Model & \bf XLM-R & \bf ModernBERT & \bf RF \\
\midrule
\rowcolor{modelrow}
\multicolumn{4}{c}{\textit{Sentence Level}} \\
GPT-5.4 & +0.020 & +0.037 & +0.087 \\
Google Translate & +0.045 & +0.060 & +0.069 \\
TranslateGemma-4B & +0.071 & +0.098 & +0.106 \\
TranslateGemma-12B & +0.062 & +0.071 & +0.115 \\
Tower-7B & +0.011 & +0.049 & +0.157 \\
\midrule
\rowcolor{modelrow}
\multicolumn{4}{c}{\textit{Document Level}} \\
GPT-5.4 & $-$0.248 & $-$0.374 & $-$0.378 \\
Google Translate & $-$0.196 & $-$0.396 & $-$0.222 \\
TranslateGemma-4B & $-$0.163 & $-$0.374 & $-$0.215 \\
TranslateGemma-12B & $-$0.204 & $-$0.396 & $-$0.193 \\
Tower-7B & $-$0.237 & $-$0.333 & $-$0.400 \\
\bottomrule
\end{tabular}
\end{table}

\subsection{Translation Shifts CEFR at Document Level}
\begin{figure}[!t]
    \centering
    \includegraphics[width=1\linewidth]{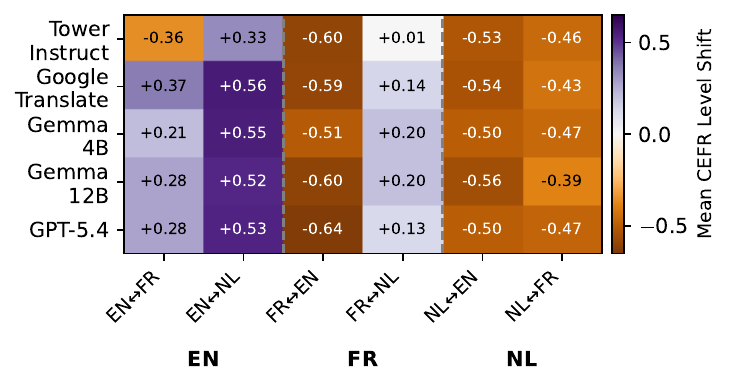}
    \caption{Mean model-anchored CEFR level shifts from back-translations across MT models for document-level texts. Level shifts are larger at the document level but remain consistent in direction across MT models for each back-translation.}
    \label{fig:task2_levelshift_document}
\end{figure}

We observe how the CEFR level of a text changes under MT by visualizing the mean model-anchored level shifts from the backtranslation process for \textsc{ComplexityMT-Preservation} (Section~\ref{subsec:preservation}) experiments across the five multilingual MT models. Figures~\ref{fig:task2_levelshift_sentence}  and~\ref{fig:task2_levelshift_document} show sentence-level and document-level results, respectively, and cross-classifier robustness checks are reported in Table~\ref{tab:threeway-delta}.

At the sentence level, the mean CEFR shift is minimal at $\Delta\ell \le +0.07$ across all MT models with 95\% confidence interval excluding zero, indicating a small but statistically distinguishable upward shift. However, this effect is not uniform across languages and is most evident with Russian, where it receives a higher CEFR classifier prediction by $\approx+0.6$ levels on average and up to $+0.95$ max, whereas using it as the source text lowers it by $\approx-0.6$ levels, down to roughly $-1.0$.

At the document level, 
we observe that \textit{the CEFR level shift is larger and consistently negative for documents} with mean $\Delta\ell$ ranging from $-0.16$ to $-0.31$ with confidence intervals excluding zero for all MT models. Considering that these selected MT models differ in their training data, language coverage, and architectures, their similar level shifts in document-level texts indicate that this is a systematic phenomenon across all models rather than an outlier related to a single model. The three-model comparison of CEFR level shifts in Table~\ref{tab:threeway-delta} further confirms this observation, where all three CEFR classifiers produce downward shifts for all MT models, with a mean $\Delta\ell$ of $-0.21$, $-0.38$, and $-0.28$ for XLM-R, ModernBERT, and the Random Forest classifiers, respectively. As discussed in the setup in Section~\ref{subsec:preservation}, these observed shifts reflect the total net effect of using backtranslation.

\subsection{Linguistic Features Expose Finer-Grained Complexity Shifts}
To systematically investigate the correlation between source text complexity and translation quality in a finer-grained setup, we design a small-scale control experiment with sentence-level texts evaluating four translation directions: EN$\rightarrow$FR, FR$\rightarrow$EN, EN$\rightarrow$RU, and RU$\rightarrow$EN. We use 200 sentence pairs per model-direction combination (3,200 instances in total), with translation quality measured via COMET. We use {\tt spaCy}\footnote{\url{https://spacy.io}} to extract 37 source text linguistic complexity features, including length statistics, POS distribution, syntactic complexity, lexical richness, and argument structures, and compute Pearson correlation coefficients against COMET.

\begin{figure}[!t]
    \centering
    \includegraphics[width=1\linewidth]{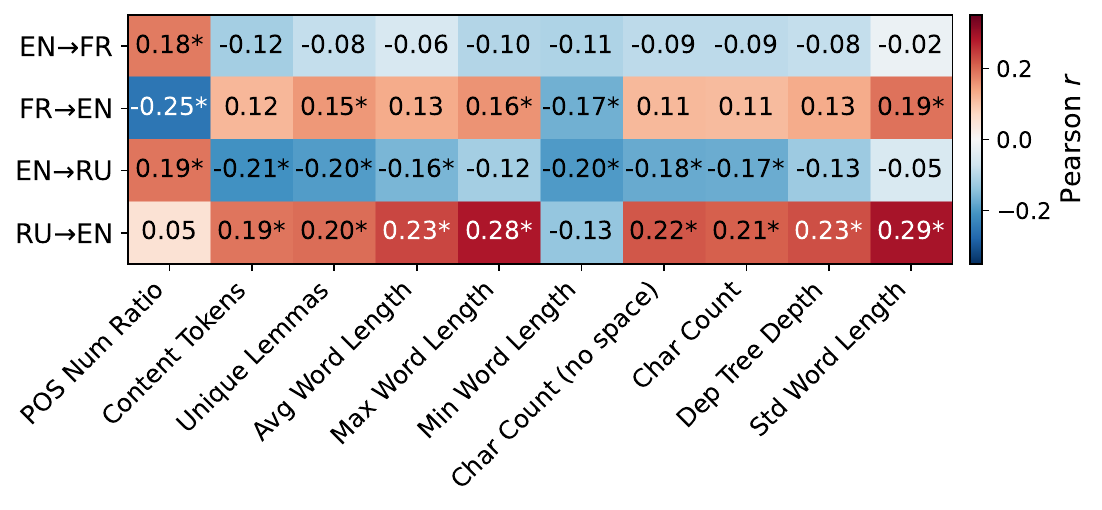}
    \caption{Pearson correlation coefficients between source-text linguistic features and COMET scores across four translation directions.}
    \label{fig:feature_comet_heatmap}
\end{figure}

Our linguistic feature analysis, as reported in Figure~\ref{fig:feature_comet_heatmap}, shows that translation directions involving Russian exhibit the strongest feature sensitivity. EN$\rightarrow$RU and RU$\rightarrow$EN display markedly opposite correlation patterns across nearly all complexity features, indicating substantial directional asymmetry. For instance, vocabulary-related features such as content tokens and unique lemmas are negatively correlated with quality in the EN$\rightarrow$RU direction ($r \approx -0.20^*$), yet positively correlated in the RU$\rightarrow$EN direction ($r \approx 0.19$--$0.20^*$). Word length features show the strongest associations in the RU$\rightarrow$EN direction (up to $r = 0.29^*$), while POS ratio yields opposing significant correlations across FR$\rightarrow$EN ($r = -0.25^*$) and EN$\rightarrow$RU ($r = 0.19^*$). These findings suggest that \textit{linguistic sensitivity is highly direction-dependent and that source-text complexity affects translation quality asymmetrically} across language pairs.

\subsection{Translation Quality Does not Predict CEFR Shift}

\begin{figure*}[!t]
    \centering
    \includegraphics[width=0.95\linewidth]{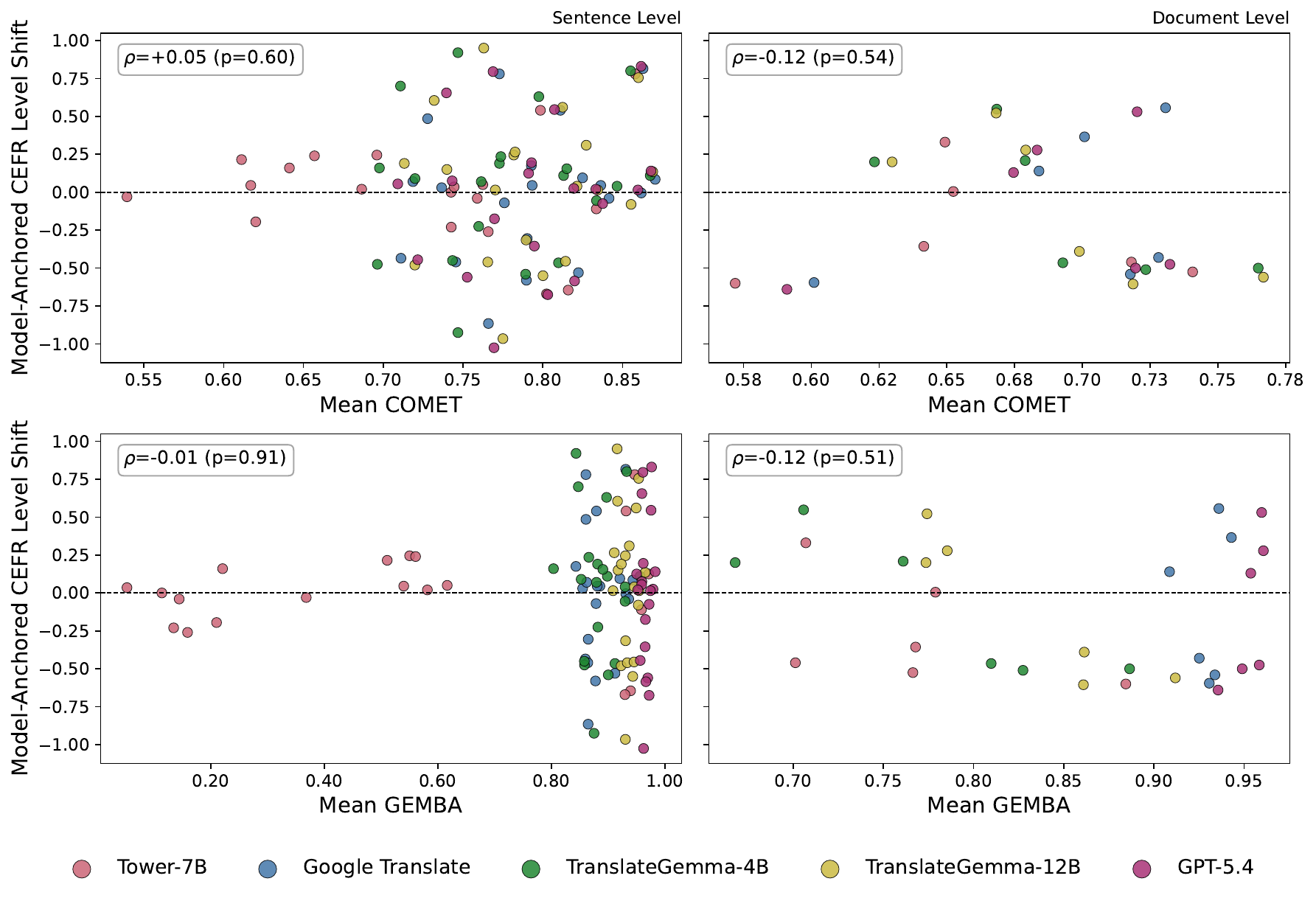}
    \caption{Scatterplot of MT quality via COMET and GEMBA versus model-anchored \ac{CEFR} level shifts for sentence and document-level texts across \ac{MT} models. The Spearman correlation is near zero in every panel ($|\rho|\le0.12$), indicating that translation quality does \textit{not} predict the magnitude or direction of its \ac{CEFR} level shift. This means that a high-quality translation is just as likely to shift the CEFR level of a text as a lower-quality translation.}
    \label{fig:cross_task_quality_vs_shift}
\end{figure*}

We visualize per-pair computed MT quality against the model-anchored CEFR level shifts in Figure~\ref{fig:cross_task_quality_vs_shift} for COMET and GEMBA, and across sentence and document level texts. We randomly sampled 100 and 30 data points for sentence- ($5$ systems $\times$ $20$ language pairs) and document-level ($5 \times 6$ language pairs), respectively.

In all four experiment runs, we obtain non-significant near-zero results for Spearman correlation ($\rho = +0.05$ and $-0.01$ at the sentence level for COMET and GEMBA, and $-0.12$ and $-0.12$ at the document level), indicating that \textit{translation quality and CEFR level shift are statistically independent}. From this, we conclude that the two subtasks under \textsc{ComplexityMT}, \textsc{ComplexityMT-Robustness} and \textsc{ComplexityMT-Preservation}, therefore measure distinct yet complementary properties of MT behavior. Since a high COMET or GEMBA does not lead to effective preservation of source CEFR levels, both a translation metric and a preservation metric may be needed for translation tasks, especially in MT applications sensitive to text complexity, such as educational content generation across multiple languages. This result demonstrates the importance of the \textsc{ComplexityMT} framework as a strong evaluation challenge for current and future MT models.

\section{Discussion}
\label{sec:disc}

\paragraph{Relationship between Machine Translation and Complexity}
Prior works have formed two distinct viewpoints on the relation of translation and complexity: complex texts are complex to translate \cite{shardlow-alva-manchego-2022-simple}, and the translation process inherently simplifies texts \cite{corpas-pastor-etal-2008-translation,wastl-etal-2025-machine}. Our empirical findings with \textsc{ComplexityMT} \textit{offer support for both viewpoints}.  \textsc{ComplexityMT-Robustness} results show that translation quality indeed declines when the source text is of a higher CEFR level, while \textsc{ComplexityMT-Preservation} results provide evidence that the simplification universal phenomenon persists at the document level. Beyond confirming both, we also showed that translation quality and complexity level shifts do not correlate with each other. Therefore, our contribution to this discourse is not to resolve any disagreement---if such a disagreement exists---but to support both aspects by showing that they are \textit{distinct dimensions of \ac{MT} behavior}.

\paragraph{Pedagogically Motivated Content Generation} 
Readability-controlled simplification is studied as a method to adapt content to different levels of text complexity, while MT enables scaling the adapted content into multiple languages, which is one of the motivations for this research. Our findings with \textsc{ComplexityMT} show that translation can shift CEFR levels differently across languages, and these results \textit{can inform content generation approaches by guiding when to simplify (at source or at target)}. For language pairs where translation tends to increase the complexity of the source text, our findings show \textit{the need for technical pipeline improvements, such as adding a readability-controlled simplification module} at the target end, and vice versa.

\section{Conclusion}
\label{sec:conclusion}
We introduced \textbf{\textsc{ComplexityMT}}, a new challenge for assessing how text complexity and machine translation interact, using the CEFR as the measure of text complexity. Across six languages and five MT systems, our experiments showed that higher \ac{CEFR} levels make translation more difficult and that MT via backtranslation lowers the target text's \ac{CEFR} level relative to the source in most languages at the document level. These two effects are also statistically independent, where translation quality does not predict the magnitude of CEFR level shifts. Together, these findings highlight the need to consider both translation quality and CEFR preservation when evaluating MT applications sensitive to text complexity, such multilingual educational content generation. 

\section*{Limitations}
\label{sec:limitations}
We identify a few limitations to our work, in terms of how we operationalize text complexity and the data used, which we discuss below. 

\paragraph{Focus on CEFR}
We use CEFR \cite{CouncilEurope2001} as the reference pedagogical construct for text complexity in our study, as it is the most widely recognized language proficiency framework across the broader education and learning community. We recognize that our findings may not directly translate to other region- or country-specific pedagogical constructs, such as the Common Core Standards (CCS) in the United States, or  China's Standards for English (CSE), to name a few. However, the methodology we followed can be replicated with other constructs, if relevant data resources are available. 

\paragraph{Use of Automatic Classifiers for CEFR}
We use three diverse, state-of-the-art CEFR classifiers from \citep{imperial-etal-2025-universalcefr}, which were trained on the massively multilingual gold-standard \textsc{UniversalCEFR} dataset. 
While automated CEFR classifiers may have inherent prediction errors, their utility in our study remains appropriate and needed, considering that our main goal includes investigating how MT models can preserve CEFR levels of texts for educational content generation, for which these automatic CEFR classifiers will be a necessary resource in the process. We also emphasize that the classifier agreement we measured and reported among our three chosen CEFR classifiers in Appendix~\ref{app:classifier_reliability} measures consistency in producing CEFR labels and should not be treated as an equal substitute for CEFR levels from human experts. Our overall findings for this paper are anchored in the relative CEFR level shifts that can be reproduced using these three classifiers, rather than relying on absolute human judgments for every language tested. This setup keeps our analyses self-contained and reproducible.


\paragraph{Data and Linguistic Coverage}
Our main results are anchored in the specific languages for which we obtained representative CEFR-labeled reference texts from \textsc{UniversalCEFR} \cite{imperial-etal-2025-universalcefr}. This includes English, Dutch, and French at the document level, and English, French, Russian, Arabic, and Hindi at the sentence level. We do not claim that our results will generalize to other languages, text formats (e.g., phrase-level), or text types (e.g., learner texts) not tested in this work. Since our sentence-level and document-level analyses cover different language test sizes and distributions, as seen in Table~\ref{tab:data_distribution}, we limit our cross-lingual findings to the languages we tested and their specific granularities. We do not claim that our findings broadly generalize to other languages we have not tested.

\paragraph{Focus on Quantitative Analysis} Our results are primarily quantitative, given the straightforward goal of empirically investigating the effect of translation on text complexity and vice-versa. We acknowledge that a well-constructed qualitative analysis would enrich and complement our work, but recognize that this can be conducted as a separate study and leave it for future work.

\section*{Acknowledgments}
This work was partly supported by the UKRI Strength in Places Fund MyWorld Project (SIPF00006/1).

JMI is supported by the National University Philippines and the UKRI Centre for Doctoral Training in Accountable, Responsible, and Transparent AI [EP/S023437/1] of the University of Bath.

ER is supported by Portuguese national funds through Fundação para a Ciência e a Tecnologia, I.P. (FCT) under projects UID/50021/2025 (DOI: \url{https://doi.org/10.54499/UID/50021/2025}), UID/PRR/50021/2025 (DOI: \url{https://doi.org/10.54499/UID/PRR/50021/2025}), and UID/04466/2025 (DOI: \url{https://doi.org/10.54499/UID/04466/2025}).

\bibliography{references}

\clearpage
\appendix
\section{Appendix}
\label{app:appendix}

\subsection{Libraries, Hyperparameters, and Configurations}
\label{app:hyperparameters}

We provide the full table of the Python libraries used and their corresponding versions in our experiments in Table~\ref{tab:libraries}. Likewise, we also provide the configurations and hyperparameter values used for the three CEFR classifiers (\textsc{XLM-R}, \textsc{ModernBERT}, and \textsc{Random Forest}) in Table~\ref{tab:classifier_configs} across all CEFR level predicitons in this study. 

\begin{table}[htbp]
\centering
\small
\begin{tabular}{ll}
\toprule
Library & Version \\
\midrule
\texttt{accelerate}        & -- \\
\texttt{bitsandbytes}      & -- \\
\texttt{huggingface-hub}   & 1.12.0 \\
\texttt{joblib}            & 1.5.3 \\
\texttt{matplotlib}        & 3.10.9 \\
\texttt{numpy}             & 2.4.4 \\
\texttt{openai}            & -- \\
\texttt{openpyxl}          & 3.1.5 \\
\texttt{pandas}            & 3.0.2 \\
\texttt{pyphen}            & 0.17.2 \\
\texttt{requests}          & 2.33.1 \\
\texttt{scikit-learn}      & 1.8.0 \\
\texttt{scipy}             & 1.17.1 \\
\texttt{sentencepiece}     & -- \\
\texttt{spacy}             & 3.8.13 \\
\texttt{spacy-stanza}      & 1.0.4 \\
\texttt{stanza}            & 1.11.1 \\
\texttt{tokenizers}        & 0.22.2 \\
\texttt{torch}             & 2.11.0 \\
\texttt{tqdm}              & 4.67.3 \\
\texttt{transformers}      & 5.6.2 \\
\texttt{tseval}            & 1.0 \\
\bottomrule
\end{tabular}
\caption{The Python libraries used in our experiments with with a main Python version of 3.14.4.}
\label{tab:libraries}
\end{table}

\begin{table}[ht]
\centering
\footnotesize
\setlength{\tabcolsep}{4pt}
\begin{tabular}{@{}>{\raggedright\arraybackslash}p{0.34\columnwidth}>{\raggedright\arraybackslash}p{0.58\columnwidth}@{}}
\toprule
Parameter & Value \\
\midrule
\multicolumn{2}{@{}l}{\textsc{XLM-R} (primary classifier)} \\
\quad Identifier   & \seqsplit{UniversalCEFR/xlm-roberta-base-cefr-all-classifier} \\
\quad Max length   & $512$ \\
\quad Truncation   & enabled \\
\quad Mode         & \texttt{model.eval()}, \texttt{torch.no\_grad()} \\
\midrule
\multicolumn{2}{@{}l}{\textsc{ModernBERT} (cross-classifier check)} \\
\quad Identifier   & \seqsplit{UniversalCEFR/ModernBERT-base-cefr-all-classifier} \\
\quad Max length   & $512$ \\
\quad Truncation   & enabled \\
\quad Mode         & \texttt{torch.inference\_mode()} \\
\midrule
\multicolumn{2}{@{}l}{\textsc{Random Forest} (feature-based)} \\
\quad Class            & \texttt{sklearn.ensemble.\seqsplit{RandomForestClassifier}} \\
\quad \texttt{n\_estimators}  & $100$ (default) \\
\quad \texttt{max\_depth}     & \texttt{None} (default) \\
\quad \texttt{random\_state}  & $42$ \\
\quad \texttt{n\_jobs}        & $-1$ \\
\quad Features         & $146$, see \cite{imperial-etal-2025-universalcefr} \\
\quad Training langs   & \texttt{en, de, fr, ar, hi, nl, ru} \\
\quad NaN handling     & impute to $0$ \\
\bottomrule
\end{tabular}
\caption{Hyperparameter and model configurations for the three CEFR classifiers (\textsc{XLM-R}, \textsc{ModernBERT}, and \textsc{Random Forest}) explored in the study.}
\label{tab:classifier_configs}
\end{table}

\begin{table}[ht]
\centering
\small
\begin{tabular}{ll}
\toprule
Parameter & Value \\
\midrule
\multicolumn{2}{l}{\textsc{COMET}} \\
\quad Checkpoint   & \texttt{Unbabel/wmt22-cometkiwi-da} \\
\quad Scale        & $[0,1]$ \\
\midrule
\multicolumn{2}{l}{\textsc{GEMBA}} \\
\quad Evaluator LLM   & GPT-5.4 \\
\quad Temperature     & $0$ \\
\quad Native scale    & $[0,100]$, rescaled to $[0,1]$ \\
\bottomrule
\end{tabular}
\caption{Configurations for the reference-free translation quality metrics COMET and GEMBA used in the study.}
\label{tab:metric_configs}
\end{table}

\subsection{Utility Prompts}
\label{app:prompts}

We provide the prompts we used for the LLM-based MT models we used in this study, including GPT-5.4, TowerInstruct-7B, TranslateGemma and for the LLM-based translation quality metric GEMBA.

\begin{promptbox}[GPT-5.4 Translation Prompt]
Translate the following \{src\_name\} sentence to \{tgt\_name\}.
Output only the translation, nothing else.\\[5pt]
\{text\}
\end{promptbox}

\begin{promptbox}[TowerInstruct-7B Translation Prompt]
Translate the following text from \{src\_name\} into \{tgt\_name\}.\\
\{src\_name\}: \{text\}\\
\{tgt\_name\}:
\end{promptbox}

\begin{promptbox}[TranslateGemma Translation Prompt]
\{"role": "user", "content": [\{\\
\ \ "type": "text",\\
\ \ "source\_lang\_code": "\{src\}",\\
\ \ "target\_lang\_code": "\{tgt\}",\\
\ \ "text": "\{text\}"\\
\}]\}
\end{promptbox}

\begin{promptbox}[GEMBA Quality Scoring Prompt]
Score the following translation from \{src\_name\} to
\{tgt\_name\} on a continuous scale from 0 to 100, where a score
of zero means ``no meaning preserved'' and a score of one hundred
means ``perfect meaning and grammar''.\\[2pt]
\{src\_name\} source: ``\{src\_text\}''\\
\{tgt\_name\} translation: ``\{tgt\_text\}''\\[5pt]
Score (0--100):
\end{promptbox}

\subsection{CEFR Classifier Reliability}
\label{app:classifier_reliability}

To investigate the robustness of the automatic CEFR classifiers we used in this work, we conduct a pairwise reliability test by computing Cohen's $\kappa$ with quadratic weights and exact-match rate on a 25\% random sample across the sentence and document-level texts across languages. We report the results in Table~\ref{tab:threeway-agreement}. Results show that all three structurally diverse CEFR classifier models exhibit moderate to high agreement ($\kappa_\text{quad}$), with XLM-R and ModernBERT showing the greatest agreement. All three CEFR classifiers also achieve $\approx 90.0+$ in adjacent agreement ($\pm1$) and  $\approx 50+$ in exact match rate.

\begin{table}[htbp]
\centering
\footnotesize
\caption{Pairwise CEFR-label agreement between XLM-R, ModernBERT, and the feature-based Random Forest model on round-trip text classification (forward and back-translations) across five MT models (25\% sample). Cohen's $\kappa$ with quadratic weights and exact-match rate.}
\label{tab:threeway-agreement}
\begin{tabular}{llrrr}
\toprule
Classifier pair & Scope & Exact & $\pm1$ & $\kappa_{\mathrm{quad}}$ \\
\midrule
\multirow{3}{*}{\shortstack[l]{XLM-R vs.\\ModernBERT}} & All & 0.662 & 0.973 & 0.835 \\
 & Sent & 0.672 & 0.977 & 0.844 \\
 & Doc & 0.610 & 0.947 & 0.695 \\
\cmidrule(lr){1-5}
\multirow{3}{*}{\shortstack[l]{XLM-R vs.\\RF}} & All & 0.508 & 0.896 & 0.584 \\
 & Sent & 0.515 & 0.896 & 0.576 \\
 & Doc & 0.469 & 0.892 & 0.480 \\
\cmidrule(lr){1-5}
\multirow{3}{*}{\shortstack[l]{ModernBERT vs.\\RF}} & All & 0.507 & 0.895 & 0.587 \\
 & Sent & 0.518 & 0.893 & 0.572 \\
 & Doc & 0.450 & 0.904 & 0.523 \\
\bottomrule
\end{tabular}
\end{table}

\subsection{Source-Anchored Level Shift}
\label{app:source_anchor}

\begin{figure*}[t]
    \centering
    \includegraphics[width=1\linewidth]{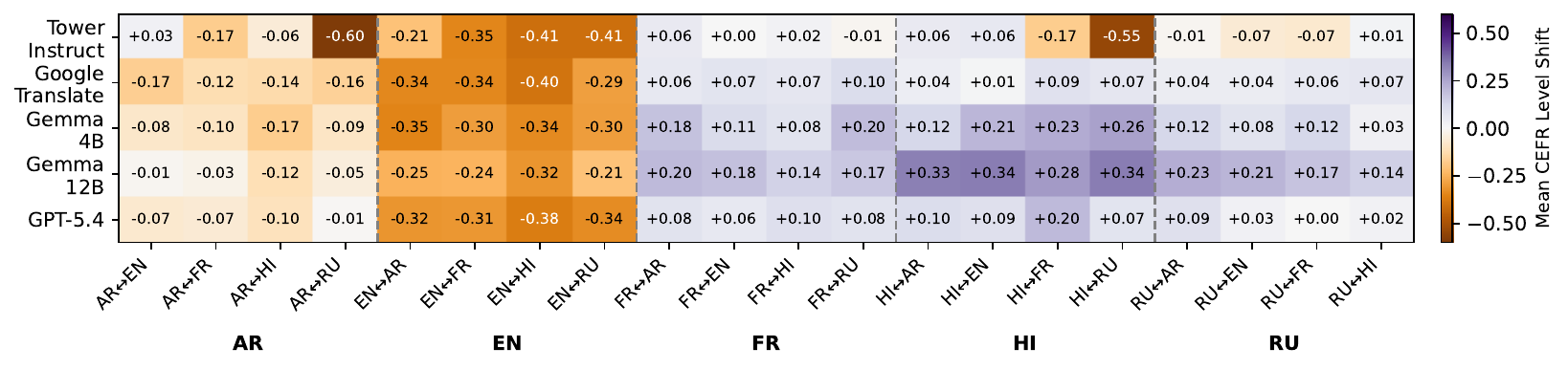}
    \caption{Mean model-anchored CEFR level shifts from back-translations across MT models for sentence-level texts. We anchor the \textsc{UniversalCEFR} classifier's source text as anchor (vs. the forward translation shown in Figure~\ref{fig:task2_levelshift_sentence}) to investigate the effects of within-language differences.}
    \label{fig:task2_levelshift_backorig_sentence}
\end{figure*}

\begin{figure*}[t]
    \centering
    \includegraphics[width=0.60\linewidth]{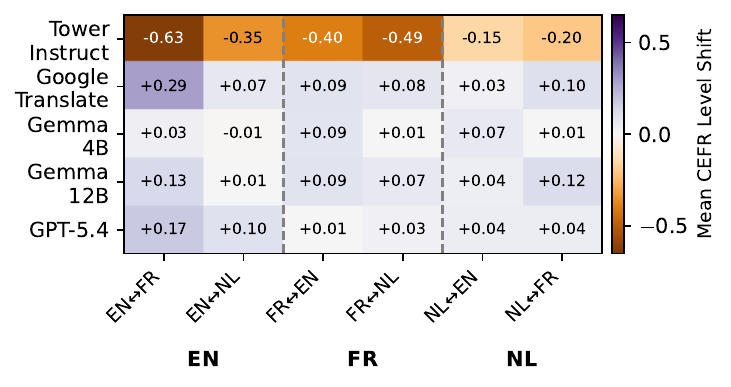}
    \caption{Mean model-anchored CEFR level shifts from back-translations across MT models for document-level texts using the source text as the anchor. Level shifts at the document level remain consistent with the sentence level of the same anchor.}
    \label{fig:task2_levelshift_backorig_document}
\end{figure*}

In Section~\ref{subsec:preservation} with \textsc{ComlexityMT-Preservation}, we discuss that we use a model-anchored shift that uses the forward and back-translation $\Delta\ell_{\text{model}} = \hat{\ell}_{\text{back}} - \hat{\ell}_{\text{fwd}}$. This represents the core translation process to quantify the CEFR level shift from the source and target languages. To investigate a within-language comparison from the back-translation process, we report a formula which is a source-anchored shift \begin{equation} \Delta\ell_{\text{source}} = \hat{\ell}_{\text{back}} - \hat{\ell}_{\text{orig}}, \end{equation} where $\hat{\ell}_{\text{orig}} = f(x)$ is the CEFR classifier's prediction on the source text. This allows measuring the CEFR level shift from the same language of the source and back-translation texts. We produce the same heatmap visualization for the $\Delta\ell_{\text{source}}$ and report the results in Figure~\ref{fig:task2_levelshift_backorig_sentence} for sentence level and Figure~\ref{fig:task2_levelshift_backorig_document} for document level.

At the sentence level, we observe that this within-language shift is relatively small. For English source texts, in particular, receive lower CEFR levels across all MT models with around $-0.21$ to $-0.41$ shifts, while French, Russian, and Hindi source texts obtain slightly higher CEFR levels compared to their original for most MT models. We do note that this within-language comparison is observable with Russian where the high sentence-level CEFR level shifts as seen in Figure~\ref{fig:task2_levelshift_sentence} using forward translation as anchor is not observed in Figure~\ref{fig:task2_levelshift_backorig_sentence} using the source text with CEFR predicted level as anchor.

At the document level, we observe a sharper pattern where Tower-7B is the only outlier model that shows consistent within-language decrease in CEFR level shifts across all backtranslations, with the highest being English to French at $-0.63$. Meanwhile, the other MT models more or less retain the source CEFR level with very minor deviations. Again, we see the distinctiveness here in using the within-language comparison, where it can cancel the classifier's per-language offset, and the backtranslation is able to return to the source text's predicted CEFR level.

\subsection{Use of LLMs}
In producing this work, we used Grammarly for minor grammar and spelling corrections, ChatGPT and Claude Code for assistance with LaTeX table and figure formatting, troubleshooting code, and issues with Matplotlib visualizations. All suggestions from these tools were scrutinized by the authors before integration into the paper. 


\end{document}